\documentclass[10pt,twocolumn,letterpaper]{article}

\usepackage[pagenumbers]{cvpr} 

\usepackage{multirow}
\usepackage{colortbl}

\definecolor{cvprblue}{rgb}{0.21,0.49,0.74}
\usepackage[pagebackref,breaklinks,colorlinks,allcolors=cvprblue]{hyperref}
\hypersetup{pdftitle={CityToolVQA: Tool-Augmented Visual Question Answering for 3D Spatial Cognition in Urban Low-Altitude Environments}, pdfauthor={Boao Yu, Yingzhen Nie, Yue Hu, Zhengqiu Zhu, Rusheng Ju}}

\def\paperID{*****} 
\def\confName{CVPR}
\def\confYear{2026}

\title{CityToolVQA: Tool-Augmented Visual Question Answering for 3D Spatial Cognition in Urban Low-Altitude Environments}

\author{Boao Yu$^{1,2}$\ \ Yingzhen Nie$^{1,2}$\ \ Yue Hu$^{1,2}$\ \ Zhengqiu Zhu$^{1,2}$\ \ Rusheng Ju$^{1,2}$\\
{\small $^{1}$College of Systems Engineering, National University of Defense Technology}\\
{\small $^{2}$National Key Laboratory of Digital Intelligent Modeling and Simulation}\\
}

\begin{document}
\maketitle
\begin{abstract}
CityToolVQA addresses the weak performance of Vision-Language Models (VLMs) on quantitative tasks in urban low-altitude visual question answering. We divide the seven tasks into qualitative and quantitative groups: qualitative questions are answered directly by the VLM, whereas quantitative questions are handled by an external visual-geometric toolchain that performs object grounding, segmentation, depth back-projection, and spatial computation. The toolchain can be attached to different VLMs in a zero-shot manner; a Depth-Assisted Prompt Inference (DAPI) fallback is triggered when the main-chain detection is invalid or unreliable, and CityToolVQA-SFT adapts the 8B backbone to tool-conditioned inputs. On the 73,324-question Open3D-VQA-v2 test set, CityToolVQA-SFT (Qwen3-VL-8B) reaches 67.6\% overall accuracy, and attaching the toolchain to ten open-source VLMs improves quantitative-task accuracy by 12.7--36.9 percentage points. These results indicate that externalizing explicit 3D geometric computation effectively complements the limited ability of RGB-only VLMs to estimate metric distances and object sizes.
\end{abstract}

\section{Introduction}
\label{sec:intro}

In tasks such as regional patrol, situational monitoring, and human--machine teaming, urban low-altitude unmanned systems need to recognize not only \emph{what} is present in a scene, but also spatial attributes of targets, such as their locations, relative positions, distances, and sizes. Visual Question Answering (VQA) takes a visual observation and a natural-language question as input, providing a natural interface through which humans can query scene information. Traditional 2D VQA mainly relies on object recognition, 2D geometric layout, and semantic cues in the image; however, when a question involves distances, directions, and scales in the physical space, semantic reasoning on the image plane alone cannot fully capture the underlying geometric relations. Three-dimensional VQA (3D-VQA) therefore extends VQA to environments with explicit spatial structure, requiring models to align language descriptions with recognized objects and spatial geometry. In urban low-altitude scenes with open object categories, large spatial-scale variation, and complex geometric relations, 3D spatial cognition and spatial-relation reasoning are essential for an agent to understand its environment.

Research on 3D-VQA and 3D language understanding has gradually moved from object recognition and localization in indoor scenes toward spatial reasoning in open urban environments. ScanQA formulates 3D question answering for RGB-D indoor scenes and achieves object-level QA by fusing 3D object candidates with linguistic representations~\cite{azuma2022scanqa}; 3D-LLM further injects 3D scene representations into large language models, enabling multiple 3D language tasks such as QA, grounding, and navigation~\cite{hong20233dllm}. Along this line, the research focus has shifted from \emph{whether a model can understand 3D scenes} to \emph{whether it can reliably reason about metric spatial quantities}. SpatialVLM substantially improves both qualitative and quantitative spatial reasoning of VLMs through training on large-scale metric-spatial data~\cite{chen2024spatialvlm}, while SpatialRGPT enhances the understanding of relative directions and distances by combining regional grounding with depth information~\cite{cheng2024spatialrgpt}. Meanwhile, CityEQA extends QA to active exploration and long-horizon embodied interaction in city spaces~\cite{zhao2025cityeqa}; Open3D-VQA further introduces seven spatial QA task types for open aerial/urban environments, and its evaluation of 13 MLLMs reveals that they perform better on relative spatial relations than on absolute distances~\cite{zhan2025open3dvqa}.

These results indicate that the key difficulty of open-urban 3D-VQA cannot be reduced to \emph{whether 3D inputs are provided to the model}. Different types of spatial questions impose distinct demands on evidence and computation: questions such as relative direction and relative size can largely be answered from 2D layout, object attributes, and language semantics, whereas inter-object distances, object-to-agent distances, and real-world sizes rely on reliable depth, camera geometry, and coordinate transformations, and their answers carry explicit physical scales. The results of SpatialVLM and SpatialRGPT show that dedicated metric information and geometric signals can improve such capabilities~\cite{chen2024spatialvlm,cheng2024spatialrgpt}; the experiments of Open3D-VQA further show that simply increasing model scale or adding 3D modalities does not guarantee stable numerical spatial reasoning; in particular, 3D LLMs fail to demonstrate significant advantages over 2D LLMs on this benchmark~\cite{zhan2025open3dvqa}. This raises a question worth further study: for 3D-VQA tasks whose spatial-evidence requirements are already distinguishable, is it necessary to keep both semantic understanding and precise geometric measurement inside a single end-to-end VLM, or should the metric computation be delegated to other means?

Tool augmentation provides a natural alternative. Toolformer enables language models to learn when and how to invoke external APIs~\cite{schick2023toolformer}; HuggingGPT employs a large language model as a controller to plan and dispatch different AI models for complex tasks~\cite{shen2023hugginggpt}; VisProg and ViperGPT translate natural-language questions into executable programs and accomplish multi-step visual reasoning by composing existing vision modules~\cite{gupta2023visual,suris2023vipergpt}, and MM-ReAct combines ChatGPT with visual experts into a multimodal reasoning system~\cite{yang2023mmreact}. ToolVQA further studies multi-step tool use in real-world visual scenes at the dataset level~\cite{yin2025toolvqa}. These works demonstrate that external specialized models and executable computation can effectively extend the capabilities of foundation models, with the main focus on tool selection, program generation, and multi-tool coordination under open-ended task conditions.

Recent studies have begun to apply this tool-augmented paradigm to spatial reasoning. SpaceTools coordinates multiple spatial-geometric tools such as depth and segmentation through double interactive reinforcement learning~\cite{chen2026spacetools}; the Geometrically-Constrained Agent (GCA) starts from the \emph{semantic--geometric gap} and constrains the tool planning and geometric computation of VLMs through formalized task constraints, making spatial reasoning more verifiable~\cite{chen2026geometric}; S-Agent combines 2D localization, 3D lifting, high-level spatial experts, and temporal memory to accumulate spatial evidence across continuous multi-view observations and videos~\cite{dai2026sagent}. In addition, DM-KG recovers entity-level 3D coordinates from street views using segmentation and metric depth, and encodes directions and distances as structured geometric priors to assist VLM reasoning~\cite{xu2026dmkg}. The work most directly related to our benchmark, MASER trains multiple modality adapters on Open3D-VQA and selects among RGB, depth, point clouds, poses, and text according to question semantics~\cite{raj2026maser}. Collectively, these works show that the spatial capability of VLMs can be enhanced through explicit geometric evidence, dedicated spatial tools, and question-conditional capability selection.

\begin{figure*}[t]
\centering
\includegraphics[width=\textwidth]{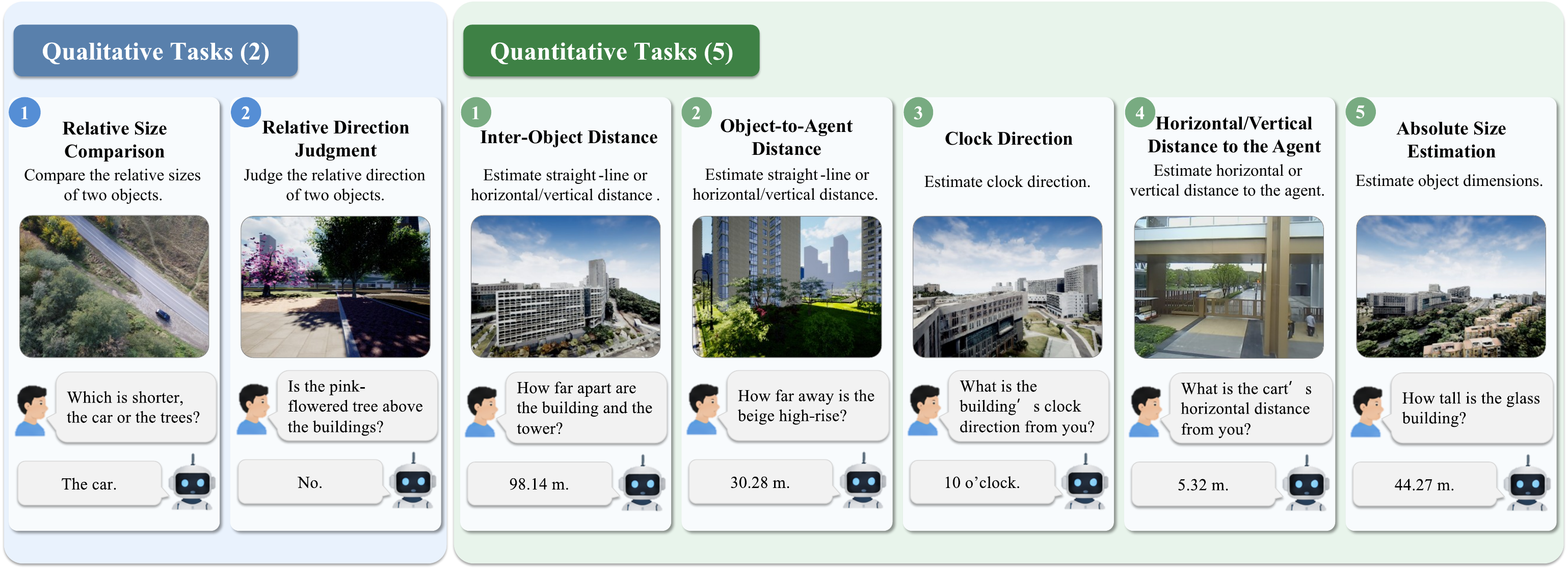}
\caption{Examples of the seven task types in Open3D-VQA-v2.}
\label{fig:tasks}
\end{figure*}

However, existing efforts mainly target open-ended tool planning, multi-tool coordination, spatiotemporal evidence accumulation, or modality selection. When task categories are predefined, the research focus can shift from \emph{how models can autonomously discover tools or modalities} to \emph{how to divide the responsibility between semantic understanding and geometric computation}. Modalities such as depth and point clouds provide spatial evidence to the VLM, yet physical quantities such as distances and sizes may still require implicit numerical inference inside the model. This leads to a more concrete question: without introducing a learnable open-ended tool planner, can we exploit the known task types to explicitly externalize the quantitative geometric computation as a unified visual-geometric alignment process, and build a \emph{plug-and-play} external toolchain in a loosely coupled, easily reusable manner? Such a toolchain should be readily plugged into various VLMs and help them perform precise geometric measurement.

Based on the above analysis, we propose CityToolVQA, whose core idea is to select the processing path according to task type. 3D-VQA tasks are divided into qualitative and quantitative categories: qualitative tasks are answered directly by the VLM, whereas for quantitative tasks the system first performs object grounding, segmentation, depth back-projection, and spatial computation, and then feeds the computation results relevant to the current question into the VLM. When the main detector fails to produce a valid or reliable target localization, the system switches to the lightweight DAPI (Depth-Assisted Prompt Inference) fallback; if DAPI still cannot obtain valid depth information, the VLM answers directly. The toolchain can be attached to different VLMs in a zero-shot manner, or further adapted to tool-conditioned inputs through CityToolVQA-SFT training.

CityToolVQA externalizes verifiable geometric computations such as distances, sizes, and directions from the VLM. Consequently, one toolchain can be reused across backbone models, and bounding boxes, segmentation masks, depth, and 3D computation results can be inspected individually, making it easier to trace errors in object detection, depth estimation, or answer generation.

Our main contributions are as follows:
\begin{itemize}
\item We propose CityToolVQA, a tool-augmented visual question answering framework for 3D spatial cognition in urban low-altitude environments. A task-type-based capability routing mechanism separates qualitative questions that VLMs handle well from quantitative questions that involve explicit geometric computation, letting the VLM and external tools collaborate with complementary capabilities.
\item We build a \emph{plug-and-play} CityToolVQA toolchain that provides verifiable 3D spatial information for quantitative questions about distances, directions, and sizes through object grounding, segmentation, depth back-projection, and spatial computation, together with a DAPI fallback mechanism that improves robustness under tool failures. Attaching the toolchain zero-shot to ten open-source VLMs brings clear gains, improving quantitative-task accuracy by 12.7--36.9 percentage points.
\item We propose CityToolVQA-SFT, a supervised fine-tuning method paired with the toolchain, which teaches the model how to answer under injected toolchain results, including tool-result understanding and task adaptation, further improving 3D-VQA capability on top of the toolchain.
\end{itemize}

\section{Problem Definition and Overall Framework}
\label{sec:problem}

\subsection{Task Definition and Capability Grouping}
\label{sec:task}

3D visual question answering aims to answer environment-related questions from 3D scene information. Given a visual observation $I$ and a natural-language question $q$, the model outputs the corresponding answer $a$; its core capabilities include object understanding, geometric attribute estimation, and spatial-relation reasoning. Open3D-VQA presents a 3D-VQA benchmark for open urban environments and further divides it into multiple spatial task types~\cite{zhan2025open3dvqa}. We use the subsequently released Open3D-VQA-v2 for our experiments, and group the seven task types into qualitative and quantitative categories according to answer form and required reasoning capability.

As shown in Fig.~\ref{fig:tasks}, Open3D-VQA-v2 contains seven task types. \emph{Relative size comparison} requires the agent to compare the sizes of two objects in the image, such as taller/shorter or wider/narrower. \emph{Relative direction judgment} requires judging the coarse directional relation between two objects, with six components: up/down/left/right/front/behind. \emph{Inter-object distance} requires estimating the straight-line distance or the horizontal/vertical projection component between two objects. \emph{Object-to-agent distance} requires estimating the straight-line distance of an object from the agent. \emph{Clock direction} requires estimating the clock direction from the agent to an object, or from one object to another. \emph{Horizontal/vertical distance to the agent} requires estimating the horizontal or vertical projection of the object-to-agent distance. \emph{Absolute size estimation} requires estimating the absolute size of a single object, such as its height or width. These seven tasks follow the task set of the original benchmark~\cite{zhan2025open3dvqa}, corresponding to the size, distance, and direction reasoning tasks under the allocentric, egocentric, allocentric-egocentric transformation, and object-centric categories. The grouping is determined by the task type annotated in the Open3D-VQA-v2 dataset and does not rely on question semantics.

We further divide the seven task types into two groups: relative size comparison and relative direction judgment are qualitative, and the remaining five are quantitative. Qualitative tasks generally ask the agent to judge whether a statement is correct or to select an answer from given options, whereas quantitative tasks all require estimating a specific distance value or a clock direction, demanding that the model recover physically meaningful spatial attributes from visual observations; they therefore pose higher spatial-reasoning requirements than qualitative semantic judgment.

\subsection{Overall Framework}
\label{sec:framework}

\begin{figure*}[t]
\centering
\includegraphics[width=\textwidth]{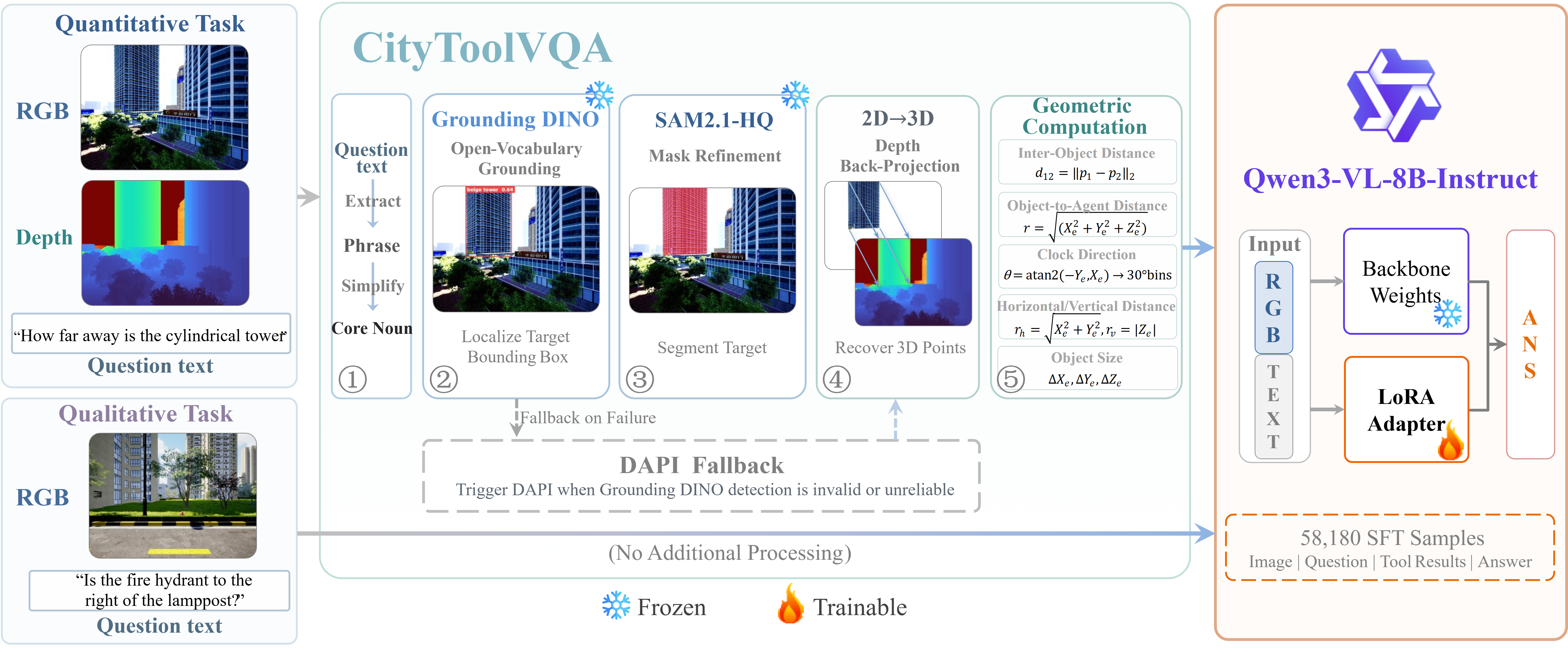}
\caption{Overall framework of CityToolVQA.}
\label{fig:framework}
\end{figure*}

CityToolVQA consists of task routing, a visual-geometric toolchain, a VLM, and a fallback module, as shown in Fig.~\ref{fig:framework}. Qualitative tasks feed the RGB image and the question directly into the VLM to generate an answer. Quantitative tasks first use Grounding DINO and SAM2.1-HQ for object grounding and segmentation, respectively, then combine the depth map to perform 3D back-projection and spatial attribute computation, and finally feed the computation results together with the original RGB image and the question text into the VLM to produce the final answer. Because VLMs do not reliably interpret depth maps, we never provide the depth map as input for either qualitative or quantitative tasks; this avoids distracting from the RGB image and the question.

After the main toolchain runs, the system inspects four classes of execution status---object detection, segmentation mask, depth data, and 3D computation---and outputs the corresponding status flags along with the tool results. The system switches to the DAPI fallback only when the main-chain detection result is invalid or unreliable, \emph{i.e.}, no valid candidate is produced, the candidate box is extremely small, or the detection confidence is very low; qualitative tasks and clock-direction tasks never trigger the fallback, and two-object questions require both objects to be localizable; the remaining status flags are used for reliability checking rather than for triggering DAPI. If DAPI still cannot obtain a valid position or depth, the system further falls back to direct VLM answering. Thus DAPI provides backup geometric measurements when the main toolchain fails to detect a target, and does not interfere with the main chain under normal conditions.

This design does not replace the VLM with tools. The VLM handles open-vocabulary understanding, question semantics, object description, and final answer generation, whereas the geometric tools compute numerical information such as distances, directions, and sizes. Task routing is determined by question type before inference, thus avoiding the additional uncertainty introduced by open-ended tool planning. Throughout this paper, the configuration without updating model parameters is denoted \emph{backbone + CityToolVQA}, and the configuration with supervised adaptation on top of Qwen3-VL-8B is denoted \emph{CityToolVQA-SFT}.

\section{The CityToolVQA Method}
\label{sec:method}

\subsection{Open-Vocabulary Grounding and Fine-Grained Segmentation}
\label{sec:grounding}

For quantitative tasks, the target description is first extracted from the question text, and Grounding DINO performs open-vocabulary object grounding~\cite{liu2024grounding}. Compared with fixed-category detectors, Grounding DINO directly accepts natural-language phrases as grounding queries, which suits urban targets with open categories and diverse descriptions such as buildings, road facilities, and vegetation. The detection stage outputs candidate boxes with confidence scores for subsequent segmentation and geometric computation.

SAM2.1-HQ, the high-quality SAM 2.1 checkpoint (\texttt{sam2.1\_hq\_hiera\_large})~\cite{ravi2025sam2,ke2023hqsam}, then performs prompt-based segmentation on the bounding boxes to obtain target masks. Compared with directly aggregating depth within a rectangular bounding box, masks suppress depth contamination from background regions such as sky, roads, and adjacent buildings. In cases involving elongated objects, partial occlusion, or multiple similar candidates, the system records the necessary detection, segmentation, and depth status flags, and passes to subsequent modules only the information relevant to the current task.

When a question involves two objects, the system localizes and segments both, and establishes the correspondence following the description order in the question. When one phrase yields multiple candidates, the candidate that best matches the description and has a valid spatial region is preferred; if the mask area is too small, valid depth is insufficient, or the candidate box falls out of bounds, an anomaly is recorded in the tool status. This design ensures that subsequent geometric computation processes only target regions with explicit image support.

\subsection{Depth Back-Projection and Spatial Computation}
\label{sec:depth}

Let $(u, v)$ be a pixel coordinate inside the target mask with valid depth $d$, and let $(f_x, f_y, c_x, c_y)$ be the camera intrinsics. Following the pinhole camera model, CityToolVQA back-projects valid pixels into the camera coordinate system:
\begin{align}
Z_c &= d, \nonumber \\
X_c &= (u - c_x)\, d / f_x, \label{eq:backproject} \\
Y_c &= (v - c_y)\, d / f_y. \nonumber \\
\intertext{To align with the agent-coordinate definition of the dataset, the camera coordinates are further converted to agent coordinates along front--left--up axes (a right-handed system):}
X_e &= Z_c, \nonumber \\
Y_e &= -X_c, \label{eq:agent} \\
Z_e &= -Y_c. \nonumber \\
\intertext{The in-mask point cloud is first filtered by valid depth and processed with robust statistics; representative positions and spatial attributes are then computed according to the question type. The straight-line, horizontal, and vertical distances from the target to the agent are}
r   &= \sqrt{X_e^2 + Y_e^2 + Z_e^2}, \nonumber \\
r_h &= \sqrt{X_e^2 + Y_e^2}, \label{eq:distance} \\
r_v &= \lvert Z_e \rvert. \nonumber
\end{align}
The inter-object distance is the Euclidean distance between the representative 3D positions of the two targets; the clock direction quantizes the azimuth $\theta=\mathrm{atan2}(-Y_e, X_e)$ in the horizontal plane into 12 positions with $30^\circ$ bins, taking the front direction as 12 o'clock and the agent's right as 3 o'clock; and the width, height, and depth of an object are estimated by the robust spatial extent of the target point cloud along the corresponding axes. All intermediate quantities of the geometric computation thus carry explicit physical meaning, which facilitates locating and reviewing anomalous measurements.

Depth statistics use the valid pixels within the mask rather than a single-point depth, and robust rules such as percentile clipping suppress edge-mixing pixels and extreme depth values. Each task uses the representative quantity consistent with its physical meaning: for object-to-agent distance, the distance between the agent and the target's representative 3D position is used; inter-object distance is computed from the difference of the two targets' representative positions; absolute size estimation is derived from the axial extent of the target point cloud.

\subsection{Task-Oriented Tool Output Organization}
\label{sec:organization}

The toolchain produces detection results, depth statistics, 3D coordinates, and various spatial attributes, but not all of them are relevant to the current question. CityToolVQA selects the necessary results by task type. For example, an inter-object-distance question receives only the 3D distance between the two targets, and an absolute-size-estimation question receives only the requested width, height, or depth. This reduces distraction caused by irrelevant numerical outputs and gives different VLMs a unified tool-output format.

Specifically, for questions such as \emph{how far is building A from the agent}, the tool context contains only the object recognition summary, the straight-line distance to the agent, and necessary reliability hints; for questions about an object's horizontal or vertical distance from the agent, such as \emph{what is the horizontal distance from target A to the agent}, only the target-matching result and the corresponding horizontal/vertical distance are kept. The tool context includes only the computation results relevant to the current question, reducing the interference of irrelevant coordinates and depth statistics on answering.

In the \emph{backbone + CityToolVQA} configuration, no additional training is performed; the toolchain provides the above results only for quantitative tasks, while qualitative tasks are still answered directly by the VLM. Therefore, changes in quantitative-task accuracy across models mainly reflect the gains brought by the toolchain.

\subsection{DAPI Depth-Assisted Fallback}
\label{sec:dapi}

\begin{figure}[t]
\centering
\includegraphics[width=\columnwidth]{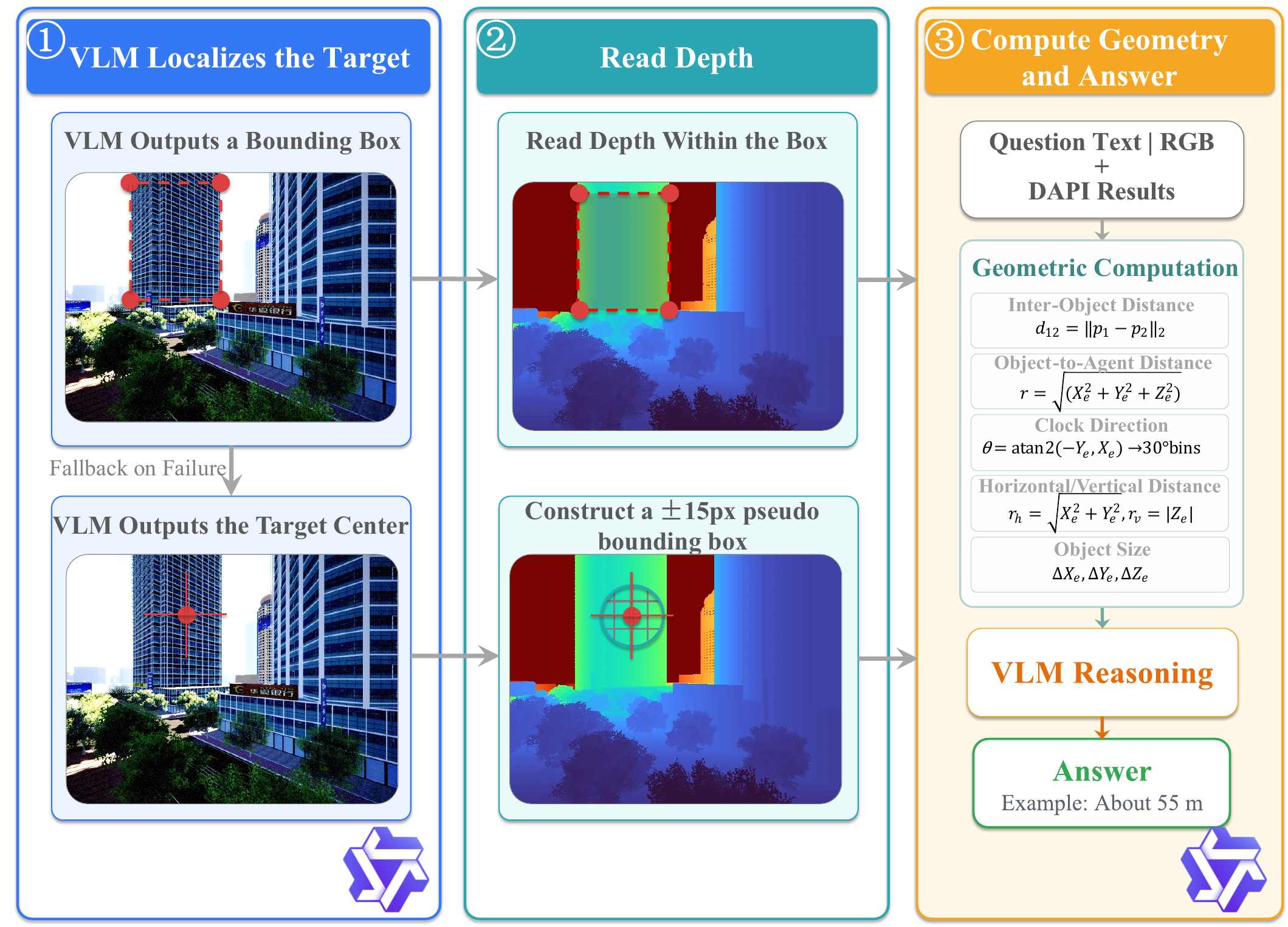}
\caption{DAPI fallback reasoning mechanism.}
\label{fig:dapi}
\end{figure}

The complete visual-geometric chain relies on the continuous success of detection, segmentation, depth, and 3D computation. To improve robustness, CityToolVQA designs Depth-Assisted Prompt Inference (DAPI) as a lightweight fallback path, as shown in Fig.~\ref{fig:dapi}. DAPI no longer depends on external detectors or segmentation modules; instead, it uses the open-vocabulary grounding capability of the VLM to obtain the target's image position, and feeds this localization into the same 3D computation process as the main chain.

When the main toolchain produces an invalid or unreliable detection (no valid candidate, an extremely small box, or very low confidence), DAPI invokes the current backbone VLM for open-vocabulary localization. It first asks the VLM to output a target bounding box and sends this localization directly into the same depth back-projection and geometric computation as the main chain, thereby skipping the faulty recognition and refinement stages. If the VLM fails to produce a valid bounding box, DAPI further asks the VLM for the target's center point and constructs a $\pm$15-pixel pseudo bounding box around it, again reusing the complete 3D computation pipeline. If DAPI still cannot form valid evidence, the system finally falls back to direct VLM answering.

CityToolVQA thus forms a three-level reasoning pipeline: \emph{complete visual-geometric toolchain $\rightarrow$ DAPI $\rightarrow$ direct VLM inference}. The complete toolchain performs the main geometric measurements; DAPI substitutes VLM localization for detection and segmentation when reliable detections are unavailable and reuses the same depth back-projection and geometric computation, with its measurement accuracy mainly constrained by the quality of VLM localization boxes; direct VLM inference is adopted only when neither of the first two levels can produce valid results.

\subsection{CityToolVQA-SFT}
\label{sec:sft}

To further adapt the model to inputs containing tool results, we perform CityToolVQA-SFT on top of Qwen3-VL-8B-Instruct~\cite{bai2025qwen3vl}. Its main role is to teach the VLM how to answer under tool-conditioned inputs, including tool-result understanding, robust answering under anomalies, and judging the credibility of tool outputs, giving the model a preliminary ability to judge \emph{when} the toolchain's measurements are trustworthy. Training samples come from the Open3D-VQA-v2 training set. To reduce the impact of tool anomalies, inconsistent coordinate conventions, and numerical outliers on the supervision signal, we first remove samples with anomalous task types or answer formats, then check the consistency of direction labels and discard numerical samples beyond preset physical ranges; in total, 8,732 samples are removed from the 66,912 training QA pairs, leaving 58,180. This procedure only constructs the training set; the validation and test sets are not involved.

Fine-tuning adopts LoRA parameter-efficient adaptation~\cite{hu2022lora} with rank 32, scaling factor 64, and dropout 0.05. Adapters are trained on 252 linear projection modules of the language backbone (36 layers $\times$ 7 projection types), with backbone weights frozen. The optimizer uses a learning rate of $5\times10^{-5}$ with cosine annealing and 70 warmup steps; training runs for 1 epoch with batch size 32, BF16 precision, and gradient checkpointing. Training takes about 3 hours on a single RTX PRO 6000 (96 GB), for a total of 1,819 steps. Since the official validation set contains a small number of duplicate questions, model selection is performed on a deduplicated subset of 10,221 validation questions; checkpoints saved during training are selected by overall accuracy on this subset, and we select the checkpoint at step 1,819.

The prompts used in SFT share the same structure as those at inference. Qualitative-task samples contain only the image and the question; quantitative-task samples contain the image, the question, and the corresponding tool context; the supervision targets are the ground-truth answers from the dataset. The training loss is computed only on target answer spans, without imposing supervision on the system prompt or the tool context. The resulting adapter retains the visual understanding capability of the backbone VLM while reliably handling input formats containing geometric tool results.

\section{Experimental Setup}
\label{sec:setup}

\subsection{Dataset and Evaluation Metrics}
\label{sec:dataset}

We use the publicly released Open3D-VQA-v2 dataset, which contains 66,912 training QA pairs, 10,282 validation QA pairs, and 73,324 test QA pairs. We strictly use the official training, validation, and test splits without re-partitioning. Training samples are constructed only from the training set, and all main experiments, cross-model experiments, and ablations are evaluated on the same test set. The test QA pairs cover simulated urban scenes from EmbodiedCity and UrbanScene, and real urban scenes from RealworldUAV and WildUAV; these four subsets contain 29,440, 40,618, 2,944, and 322 QA pairs, respectively. We neither re-partition the test set nor filter questions by model performance; all compared methods use the same question order and reference answers.

Grading rules correspond to the answer form of each task. Numerical questions adopt a 25\% relative error tolerance: a prediction is correct if its relative error from the reference is no greater than 0.25. Clock-direction questions allow a cyclic clock-distance tolerance of one hour. The remaining open-text answers are judged by a locally deployed DeepSeek-R1-Distill-Qwen-14B semantic judge that decides whether the prediction and the reference answer are equivalent~\cite{guo2025deepseek}. All models use temperature 0, and overall accuracy is computed directly over all 73,324 QA pairs, so tasks with larger sample sizes carry higher weight.

\subsection{Baselines and Implementation}
\label{sec:baselines}

The main comparison covers open-source VLMs including Qwen3-VL-4B/8B/30B-A3B/32B, Qwen3.6-27B, Qwen3.8-27B, InternVL3.5-8B/38B, Kimi-VL-A3B, and ERNIE-4.5-VL-28B-A3B~\cite{bai2025qwen3vl,qwen2026qwen36,qwen2026qwen38,wang2025internvl35,kimi2025kimivl,ernie2025technical}. Except for CityToolVQA-SFT, all backbones are evaluated without task-specific fine-tuning; the RGB-only setting is defined below as the direct-inference baseline.

We refer to the setting that takes only the RGB image and the question, without the CityToolVQA toolchain, as the \emph{VLM direct-inference baseline}. To keep comparison conditions identical, the direct-inference baseline uses unified image--question prompts and output requirements; the \emph{+CityToolVQA} configuration keeps the same backbone model and decoding settings, adding the CityToolVQA tool context only for quantitative tasks, without updating model parameters. CityToolVQA additionally uses the depth map paired with each sample and performs explicit geometric computation; the corresponding comparisons therefore evaluate the gains from structured 3D information and explicit computation rather than the direct superiority of models with different parameter scales. DAPI is triggered only by invalid or unreliable main-chain detection; direct VLM answering is used only if the fallback also fails to produce valid geometric evidence.

\subsection{Reproducibility and Comparison Protocol}
\label{sec:fairness}

Cross-model comparisons adopt unified prompt templates, the same maximum output budget, and deterministic decoding. The parameters, task definitions, and interfaces of the CityToolVQA toolchain remain identical across backbones; detection thresholds and prompt templates are not re-optimized for any single backbone, and the \emph{+CityToolVQA} settings in generalization experiments all use the same visual-geometric tool interface. For models that support a thinking mode, explicit thinking output is disabled at inference so that the answering protocol stays consistent with short-answer VQA.

Main experiments and ablations are all run on the complete 73,324 test QA pairs and aggregated by a unified grading program. Inference outputs go through consistent format parsing and validity checks; anomalies such as empty outputs, parsing failures, and truncation are handled by preset rules, ensuring identical evaluation protocols across models.

\section{Experimental Results and Analysis}
\label{sec:results}

\begin{table*}[t]
  \centering
  \caption{Accuracy of different methods on the seven Open3D-VQA-v2 tasks (\%).}
  \label{tab:main}
  \scriptsize
  \begin{tabular}{lcccccccc}
    \toprule
    \multirow{2}{*}{Method} & \multicolumn{2}{c}{Qualitative} & \multicolumn{5}{c}{Quantitative} & \multirow{2}{*}{Overall} \\
    \cmidrule(lr){2-3} \cmidrule(lr){4-8}
     & Size Comp. & Direction & Inter-Dis & Ego-Dis & Clock & H/V-Dis & Size Est. & \\
    \midrule
    Qwen3-VL-4B        & 57.4 & 67.2 & 7.0  & 5.3  & 34.3 & 8.3  & 9.8  & 53.7 \\
    InternVL3.5-8B     & 60.1 & 65.6 & 8.3  & 13.2 & 27.1 & 7.8  & 12.8 & 53.3 \\
    Qwen3-VL-8B        & 57.6 & 69.0 & 9.3  & 6.4  & 42.4 & 7.8  & 14.4 & 55.6 \\
    Kimi-VL-A3B        & 55.2 & 59.9 & 4.0  & 2.8  & 14.7 & 4.9  & 6.0  & 47.5 \\
    Qwen3.6-27B        & 59.3 & 71.1 & 7.3  & 9.6  & 56.1 & 8.3  & 9.5  & 57.5 \\
    Qwen3.8-27B        & 59.5 & 69.8 & 6.7  & 11.4 & 52.9 & 9.9  & 8.9  & 56.6 \\
    ERNIE-4.5-VL-28B-A3B & 57.3 & 62.8 & 9.3 & 11.9 & 31.5 & 12.5 & 11.7 & 51.5 \\
    Qwen3-VL-30B-A3B   & 57.4 & 70.1 & 11.0 & 6.8  & 44.7 & 9.7  & 14.1 & 56.5 \\
    Qwen3-VL-32B       & 59.2 & 69.8 & 10.6 & 12.1 & 47.6 & 10.5 & 11.2 & 56.7 \\
    InternVL3.5-38B    & 59.8 & 66.0 & 9.0  & 18.1 & 48.9 & 14.5 & 9.4  & 54.7 \\
    \rowcolor{gray!15}
    CityToolVQA-SFT (Qwen3-VL-8B) & 65.4 & 75.2 & 27.5 & 65.7 & 74.1 & 52.2 & 39.7 & 67.6 \\
    \bottomrule
  \end{tabular}
  \\[2pt]
  \parbox{0.95\textwidth}{\scriptsize Note: Size Comp. = relative size comparison; Direction = relative direction judgment; Inter-Dis = inter-object distance; Ego-Dis = object-to-agent distance; Clock = clock direction; H/V-Dis = horizontal/vertical distance to the agent; Size Est. = absolute size estimation. Sample sizes of the seven tasks are 12{,}752, 44{,}632, 4{,}782, 1{,}594, 3{,}188, 3{,}188, and 3{,}188, totaling 73{,}324 QA pairs.}
\end{table*}

\begin{table*}[t]
  \centering
  \caption{Zero-shot gains of attaching CityToolVQA to different backbone VLMs.}
  \label{tab:zeroshot}
  \scriptsize
  \begin{tabular}{lcccccccc}
    \toprule
    \multirow{2}{*}{Backbone} & \multicolumn{3}{c}{VLM direct inference (\%)} & \multicolumn{3}{c}{+CityToolVQA (\%)} & \multicolumn{2}{c}{Gain $\Delta$ (pp)} \\
    \cmidrule(lr){2-4} \cmidrule(lr){5-7} \cmidrule(lr){8-9}
     & Qual. & Quant. & Overall & Qual. & Quant. & Overall & $\Delta$Quant. & $\Delta$Overall \\
    \midrule
    Qwen3-VL-4B          & 65.0 & 13.1 & 53.7 & 65.0 & 46.2 & 60.9 & +33.1 & +7.2 \\
    InternVL3.5-8B       & 64.4 & 13.4 & 53.3 & 64.4 & 26.1 & 56.1 & +12.7 & +2.8 \\
    Qwen3-VL-8B          & 66.5 & 16.3 & 55.6 & 66.5 & 46.7 & 62.2 & +30.4 & +6.6 \\
    Kimi-VL-A3B          & 58.8 & 6.6  & 47.5 & 58.9 & 43.5 & 55.6 & +36.9 & +8.1 \\
    Qwen3.6-27B          & 68.4 & 17.9 & 57.5 & 68.4 & 46.3 & 63.6 & +28.4 & +6.1 \\
    Qwen3.8-27B          & 67.4 & 17.5 & 56.6 & 67.4 & 43.9 & 62.3 & +26.4 & +5.7 \\
    ERNIE-4.5-VL-28B-A3B & 61.6 & 15.1 & 51.5 & 61.8 & 40.2 & 57.1 & +25.1 & +5.6 \\
    Qwen3-VL-30B-A3B     & 67.3 & 17.7 & 56.5 & 67.4 & 47.2 & 63.0 & +29.5 & +6.5 \\
    Qwen3-VL-32B         & 67.4 & 18.2 & 56.7 & 67.5 & 46.6 & 63.0 & +28.4 & +6.3 \\
    InternVL3.5-38B      & 64.6 & 19.1 & 54.7 & 64.7 & 44.2 & 60.2 & +25.1 & +5.5 \\
    \bottomrule
  \end{tabular}
  \\[2pt]
  \parbox{0.95\textwidth}{\scriptsize Note: every setting is evaluated on the full test set of 73{,}324 QA pairs; backbone parameters are frozen when attaching CityToolVQA. pp denotes percentage points.}
\end{table*}

\begin{table}[t]
  \centering
  \caption{Component ablation results of CityToolVQA.}
  \label{tab:ablation}
  \scriptsize
  \setlength{\tabcolsep}{3pt}
  \begin{tabular}{cccccccrr}
    \toprule
    G-DINO & SAM & Num. & DAPI & SFT &
    Qual. & Quant. & Overall & $\Delta$ \\
     & & & & & (\%) & (\%) & (\%) & (pp) \\
    \midrule
    \checkmark & \checkmark & \checkmark & \checkmark & \checkmark & 73.0 & 48.0 & 67.6 & --- \\
    $\times$   & \checkmark & \checkmark & \checkmark & \checkmark & 73.0 & 30.8 & 63.8 & $-$3.8 \\
    \checkmark & $\times$   & \checkmark & \checkmark & \checkmark & 72.9 & 39.5 & 65.7 & $-$1.9 \\
    \checkmark & \checkmark & $\times$   & \checkmark & \checkmark & 73.3 & 26.9 & 63.2 & $-$4.4 \\
    \checkmark & \checkmark & \checkmark & $\times$   & \checkmark & 72.9 & 47.9 & 67.5 & $-$0.1 \\
    $\times$   & $\times$   & $\times$   & \checkmark & \checkmark & 73.0 & 26.7 & 62.9 & $-$4.7 \\
    $\times$   & $\times$   & $\times$   & $\times$   & \checkmark & 72.9 & 25.2 & 62.5 & $-$5.1 \\
    $\times$   & $\times$   & $\times$   & $\times$   & $\times$   & 66.5 & 16.3 & 55.6 & $-$12.0 \\
    \bottomrule
  \end{tabular}
  \\[2pt]
  \parbox{0.95\columnwidth}{\scriptsize Note: evaluated on the full test set of 73{,}324 QA pairs; pp denotes percentage points. G-DINO = Grounding DINO, SAM = SAM2.1-HQ, Num.\ = numeric computation, SFT = supervised fine-tuning.}
\end{table}

\subsection{Main Comparison}
\label{sec:main}

Table~\ref{tab:main} presents the results of ten VLM direct-inference baselines and CityToolVQA-SFT on the seven task types. The overall accuracy of the VLM baselines ranges from 47.5\% to 57.5\%, with Qwen3.6-27B the highest at 57.5\%. Larger VLMs already show an advantage on relative direction judgment and clock direction, but inter-object distance, object-to-agent distance, and absolute size estimation remain at low levels, indicating that simply increasing model scale does not by itself solve quantitative problems such as distances and sizes.

CityToolVQA-SFT reaches 67.6\% overall accuracy, 12.0 percentage points above the Qwen3-VL-8B direct-inference baseline. The gain comes mainly from quantitative tasks: object-to-agent distance rises from 6.4\% to 65.7\%, horizontal/vertical distance to the agent from 7.8\% to 52.2\%, and absolute size estimation from 14.4\% to 39.7\%. Relative direction judgment also improves from 69.0\% to 75.2\%, showing that SFT consistently strengthens both qualitative and quantitative performance and helps the model better exploit the toolchain's measurements on quantitative tasks.

Across parameter scales of the VLM baselines, enlarging the model improves some clock-direction and relative-direction results, but distance and size tasks remain weak. Qwen3.6-27B and Qwen3.8-27B reach 56.1\% and 52.9\% on clock direction, higher than several smaller models, yet their inter-object distance is only 7.3\% and 6.7\%, and absolute size estimation only 9.5\% and 8.9\%. InternVL3.5-38B reaches 18.1\% on object-to-agent distance, still far below the 65.7\% of CityToolVQA-SFT. These results indicate that larger VLMs already possess certain capabilities on tasks with discrete answers such as relative direction and clock direction, whereas object distances, agent distances, and size estimation stay at low accuracy: VLMs can exploit the 2D geometric layout in images for partial spatial reasoning, but remain unreliable at recovering metric scale from RGB inputs.

\subsection{Zero-Shot Attachment of CityToolVQA}
\label{sec:zeroshot}

To disentangle the effect of the toolchain itself from supervised fine-tuning, Table~\ref{tab:zeroshot} compares direct inference and CityToolVQA attachment for ten VLMs under zero-shot conditions. Overall accuracy improves by 2.8--8.1 percentage points for all ten models, and quantitative-task accuracy improves by 12.7--36.9 percentage points, showing that the toolchain benefits VLMs of different scales in a zero-shot manner. Since the CityToolVQA toolchain provides measurements only for quantitative tasks that require explicit geometric computation, while qualitative tasks keep the original VLM inference path, the qualitative accuracy of each model in Table~\ref{tab:zeroshot} stays nearly unchanged before and after attachment.

Taking Qwen3-VL-8B as an example, after attaching CityToolVQA, quantitative-task accuracy rises from 16.3\% to 46.7\%, with the gain mainly from the explicit spatial computation provided by the toolchain; combined with Table~\ref{tab:ablation}, further SFT lifts quantitative accuracy modestly to 48.0\%. The toolchain's improvement on quantitative tasks thus does not depend on the subsequent SFT stage, and the backbones can already interpret the toolchain's geometric measurements well without task-specific fine-tuning.

The quantitative gains are highly consistent across models: Kimi-VL-A3B improves by 36.9 percentage points, the largest gain; Qwen3-VL-30B-A3B and Qwen3-VL-32B improve by 29.5 and 28.4 percentage points; InternVL3.5-38B and ERNIE-4.5-VL-28B-A3B both improve by 25.1 percentage points. CityToolVQA thus provides effective geometric measurements on all tested VLMs. The weakest quantitative baseline, Kimi-VL-A3B, obtains the largest gain, whereas InternVL3.5-8B gains relatively little (12.7 percentage points), and its post-attachment quantitative accuracy (26.1\%) is also the lowest among all models; we conjecture this relates to how well the model follows the textual format of tool evidence.

\subsection{Ablation Study}
\label{sec:ablation}

To examine the role of key components in the geometric toolchain, we conduct five component ablation variants plus an SFT-only reference with Qwen3-VL-8B on the full test set. Unlike removing modules only at inference, each ablation configuration reconstructs the training inputs according to the available tools and is retrained under the same SFT budget, controlling the train--inference distribution gap. When Grounding DINO is removed, quantitative samples entering the toolchain are localized by VLM-output bounding boxes as in DAPI. When SAM2.1-HQ is removed, the Grounding DINO bounding boxes are kept without refinement, and depth back-projection and numeric computation are performed directly on them. When numeric computation is removed, localization information such as detection boxes, center points, and confidence scores is kept, while all computed quantities---distances, sizes, depth, 3D coordinates, and clock positions---are removed from the tool evidence. When DAPI is removed, neither training nor testing contains fallback evidence, and samples failing the main chain are answered directly by the VLM. The DAPI-localization-only configuration removes main-chain detection, segmentation, and numeric computation, so quantitative samples receive only the VLM localization hint.

As shown in Table~\ref{tab:ablation}, the complete CityToolVQA-SFT achieves 48.0\% quantitative accuracy, whereas SFT alone without the toolchain reaches only 25.2\%, a gap of 22.8 percentage points. Removing Grounding DINO, SAM2.1-HQ, or the numeric computation module drops quantitative accuracy to 30.8\%, 39.5\%, and 26.9\%, respectively, while the qualitative results of all SFT groups remain stable at 72.9\%--73.3\%. The complete toolchain is thus most critical for quantitative performance, whereas qualitative performance is mainly carried by the VLM itself and the subsequent SFT, consistent with the design that the toolchain does not alter the processing path of qualitative tasks.

All ablation configurations still outperform the SFT-only group without the toolchain in overall accuracy, but remain below the full method. Among the three single-component removals, numeric computation causes the largest drop, indicating that localization information alone cannot replace explicit numerical computation. After removing Grounding DINO, localization is taken over by VLM-output bounding boxes, and quantitative accuracy falls from 48.0\% to 30.8\%, showing that VLM localization accuracy is still clearly lower than that of dedicated detectors. Removing SAM2.1-HQ causes the smallest overall drop, but background depth inside bounding boxes more easily contaminates the point-cloud statistics. Removing the DAPI fallback only lowers overall accuracy by 0.1 percentage points: the DAPI trigger rate in the full toolchain is merely 0.022\%, \emph{i.e.}, only 16 of the 73,324 test samples. On these 16 samples, however, DAPI raises the paired accuracy from 6.3\% to 37.5\%; its value lies in rescuing samples where the main chain fails. The DAPI-localization-only configuration drops overall accuracy by 4.7 percentage points, comparable to removing numeric computation, indicating that the quantitative gains come mainly from explicit geometric computation rather than the localization hint itself. These results show that the toolchain components are complementary, and the overall gain is not produced by any single module.

\subsection{Discussion and Limitations}
\label{sec:discussion}

CityToolVQA-SFT and the VLM direct-inference baselines use different information. Larger VLMs already reason well on relative direction and some clock-direction questions from RGB inputs alone, whereas CityToolVQA-SFT additionally exploits depth maps and explicit geometric computation, so its advantage concentrates on quantitative tasks such as distances and sizes. The comparison here analyzes the performance change brought by structured 3D information and geometric computation, rather than claiming that the 8B model surpasses larger VLMs in parameter scale.

CityToolVQA supports two complementary deployment modes. The zero-shot \emph{+CityToolVQA} mode requires no additional training and suits rapid migration to new VLM backbones; CityToolVQA-SFT requires an additional adaptation stage targeted at the tool context, further improving qualitative performance and helping the model use tool results more reliably and follow the short-answer output format. DAPI, shared by both modes as an inference-time fallback, handles only explicit execution failures of the complete toolchain, and its role is independent of the model adaptation of SFT.

CityToolVQA still depends on object grounding quality, segmentation quality, depth-map reliability, and camera parameters. Ambiguous open-vocabulary target descriptions, occlusion, tiny objects, and depth discontinuities may propagate errors into subsequent 3D computation. Moreover, the toolchain provides explicit spatial evidence rather than exact, always-reliable ground-truth answers; on hard samples the toolchain's measurements can still be inaccurate, and the model still needs to make holistic judgments combining task semantics and tool-output quality. Although DAPI bypasses external detection and segmentation and reuses the same geometric computation, its measurement accuracy depends on the boundary quality of VLM open-vocabulary localization and remains below the full main chain equipped with detection and refinement. The current task routing uses deterministic rules and assumes the task type is known before inference. Future work includes multi-candidate geometric estimation, cross-frame and multi-view fusion, and adaptive toolchains that classify task categories by question semantics when the question type is unknown.

\section{Conclusion}
\label{sec:conclusion}

To address the substantial performance gap between quantitative and qualitative tasks in urban low-altitude visual question answering, we proposed CityToolVQA. The method selects the processing path between direct VLM answering and the visual-geometric toolchain according to task type, invokes the DAPI fallback when main-chain target detection is invalid or unreliable; CityToolVQA-SFT further adapts Qwen3-VL-8B to tool-conditioned inputs. On the complete Open3D-VQA-v2 test set, CityToolVQA-SFT reaches 67.6\% overall accuracy; when CityToolVQA is attached to ten open-source VLMs in a zero-shot manner, quantitative-task accuracy improves by 12.7--36.9 percentage points. Ablations show that object grounding, fine-grained segmentation, and explicit numeric computation all contribute to quantitative performance. These results demonstrate that explicit 3D computation can effectively complement the limited ability of RGB-only VLMs to estimate metric distances and object sizes.

{
    \small
    \bibliographystyle{ieeenat_fullname}
    \bibliography{refs}
}

\end{document}